# Comparing Baseline Shapley and Integrated Gradients for Local Explanation: Some Additional Insights[1]

Tianshu Feng, Zhipu Zhou[2], Joshi Tarun, and Vijayan N. Nair

Advanced Technologies for Modeling
Corporate Model Risk, Wells Fargo

**Abstract**

There are many different methods in the literature for local explanation of machine learning results. However, the methods differ in their approaches and often do not provide same explanations. In this paper, we consider two recent methods: Integrated Gradients (Sundararajan, Taly, & Yan, 2017) and Baseline Shapley (Sundararajan and Najmi, 2020). The original authors have already studied the axiomatic properties of the two methods and provided some comparisons. Our work provides some additional insights on their comparative behavior for tabular data. We discuss common situations where the two provide identical explanations and where they differ. We also use simulation studies to examine the differences when neural networks with ReLU activation function is used to fit the models.

**Keywords**: Attribution methods, Local feature importance, Path integrals

## 1 INTRODUCTION

Consider a fitted model $f(x)$ using a complex machine learning (ML) algorithm with $P$ predictors $x = (x_1, \ldots, x_P)$. There are many methods in the literature to explain the results of the fitted model. These are typically referred to as post hoc methods and can be classified into global or local techniques. Within local explainaibility methods, one class aims to decompose the difference between the fitted $f(x)$ model at two different points: $x^O$, a point of interest and $x^R$, a reference point, and attributes it to the different variables $\{x_1, \ldots, x_P\}$. As such, they are also called attribution methods.

Many methods have been proposed for local attribution. They include:

- Shapley-based methods: SHAP (SHapley Additive exPlanation) (Lundberg & Lee, 2017), Kernel SHAP (Lundberg & Lee, 2017), Tree SHAP (Lundberg, Erion, & Lee, 2019; Lundberg S. M., et al., 2019), Deep SHAP (Ancona, Oztireli, & Gross, 2019; Chen, Lundberg, & Lee, 2019; Ancona, Oztireli, & Gross, 2019; Ancona, Ceolini, Oztireli, & Gross, 2018); and Baseline SHAP (Sundararajan & Najmi, 2020).
- Deep LIFT (Lundberg & Lee, 2016; Shrikumar, Greenside, & Kundaje, 2019; Shrikumar, Greenside, Shcherbina, & Kundaje, 2016), and
- Integrated gradients or IG: (Shrikumar, Greenside, & Kundaje, 2019; Ancona, Ceolini, Oztireli, & Gross, 2018; Merrill, Ward, Kamkar, Budzik, & Merrill, 2019; Jha A. , Aicher, Gazzara, Singh, & Barash, 2020; Sundararajan, Taly, & Yan, 2017). IG is also based on the Shapely concept and is a generalization of the ideas in (Aumann & Shapley, 2015).

---

[1] The views expressed in the paper are those of the authors and do not represent the views of Wells Fargo.
[2] Corresponding author (email: Lucas.Zhou@wellsfargo.com)

It is known that different approaches can lead to different local explanations, so a choice among them often depends on computational simplicity and user preference. We focus here on two recent methods called Baseline Shapley (BShap) and Integrated Gradients (IG) that are both easy to compute and are based on path integrals (to be explained in the next section). The papers that introduced these techniques (Sundararajan, Taly, & Yan, 2017; Sundararajan & Najmi, 2020) have studied their axiomatic properties and also provided some theoretical comparisons. See also other references listed above.

The present paper supplements results in the literature by examining their performance under common statistical models for tabular data, and discusses when the attributions are the same and when they are different. In addition, the paper compares the differences for the true underlying models versus differences when they are fitted using neural networks with the commonly used ReLU activation function. IG is not applicable with tree-based models as their response surface is not differentiable.

The rest of this paper is structured as follows. Section 2 reviews IG and BShap as tools for local model diagnostics. Section 3 discusses commonly used models where results of BShap and IG are the same and where they are different. Section 4 describes results from a simulation study to compare their differences when the models are fitted using a neural network with ReLU activation. Section 4 describes results from simulation studies.

## 2 METHODOLOGY

### 2.1. Baseline SHAP

Baseline Shap (BShap), recently introduced by (Sundararajan & Najmi, 2020), is one version of Shapley-based approaches for global and local interpretation of ML algorithms. The Shapley concept itself was originally proposed by (Shapley, 1951) in the context of cooperative game theory.

To define BShap, recall that $x = (x_1, \ldots, x_P)$ is the set predictors, and $f(x)$ is the fitted model. Let $x^O$ be the point of interest where the attribution is to be made, and $x^R$ be the reference point. Further, let $\mathbf{P}$ denote the set $\{1, \ldots, P\}$, $\mathbf{S} \subseteq \mathbf{P}$ be a subset, and $|\mathbf{S}|$ denote its cardinality. Then, BShap for the $i$-th predictor is given by

$$\phi_i^{\text{SHAP}}(x^O; x^R) = \sum_{S \subseteq P \setminus \{i\}} \frac{|S|!\,(P - |S| - 1)!}{P!} \big(v(S \cup \{i\}) - v(S)\big), \tag{1}$$

where $v(S) = f(x_S^O; x_{P \setminus S}^R)$. Note that $\phi_i^{\text{SHAP}}$ depends on $f$ through $v(\cdot)$.

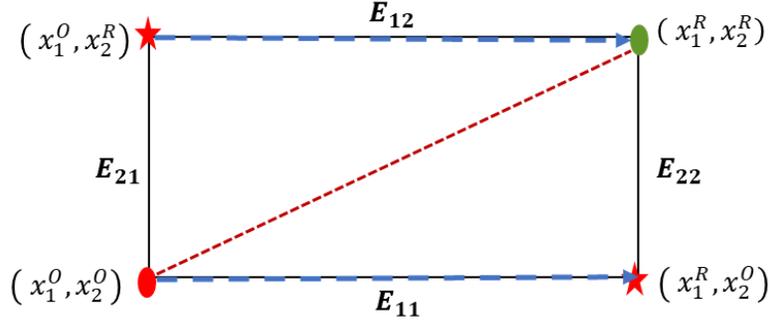

Figure 1: Illustration of the difference between $\phi_1^{\text{SHAP}}(x^O; x^R)$ and $\phi_1^{IG}(x^O; x^R)$

To make the expression in Eq (1) concrete, consider the simple two-dimensional case in Figure 1. In this case, $P = \{1, 2\}$. If $S = \{1\}$, $P \setminus S = \{2\}$, so, $v(S) = f(x_S,; x_{P \setminus S}^R) = f(x_1^O, x_2^R)$. Now, looking at all possible subsets $S$, and substituting the corresponding values for $v(S)$, we can get

$$\phi_1^{\text{SHAP}}(x^O; x^R) = \frac{1}{2}(E_{11} + E_{12}),$$

where $E_{11} = [f(x_1^O, x_2^O) - f(x_1^R, x_2^O)]$, and $E_{12} = [f(x_1^O, x_2^R) - f(x_1^R, x_2^R)]$. Thus, $\phi_1^{\text{SHAP}}$ is the average of the path integrals of the two blue dashed lines in Figure 1. See Nair et al. (2022) for illustrations in the three-dimensional case as well application of BShap to explain adverse actions in credit loan decisions. See also Nair et al. (2022) for exact computations of BShap in models with lower-order interactions. Castro, Gomez, & Tejada (2009) provide a sampling-based approach to computing BShap that is useful in high-dimensional problems.

## 2.2. Integrated gradients

Integrated gradients (IGs) is another recent local attribution method, and it is widely used in computer vision and deep learning (Sundararajan, Taly, & Yan, 2017). It is related to the Aumann-Shapley method (Aumann & Shapley, 2015), which is an extension of discrete Shapley values to continuous settings (Sundararajan & Najmi, 2020).

Let $f$ be a continuous and almost everywhere (a.e.) differentiable function. Then, $\phi_i^{IG}(x^O, x^R)$ is defined as the integral of a suitable gradient of $f$ along the straight-line path between $x^O$ and $x^R$. Formally, define the path between $x^O$ and $x^R$ as $\gamma(\alpha) = (\alpha x^O + (1 - \alpha) x^R)$ for $\alpha \in [0, 1]$. Then, IG for the $i$-th feature $x_i$ is defined as:

$$\phi_i^{IG}(x^O, x^R) = \int_{\alpha=0}^{1} \frac{\partial f(\gamma(\alpha))}{\partial \gamma_i(\alpha)} \frac{\partial \gamma_i(\alpha)}{\partial \alpha} d\alpha$$
$$= (x_i^O - x_i^R) \int_{\alpha=0}^{1} \frac{\partial f((x^R + \alpha(x^O - x^R)))}{\partial x_i} d\alpha.$$

In Figure 1, $\phi_i^{IG}(x^O, x^R)$ is the path integral of the partial derivative of $f$ with respect to $x_1$ along the <u>dashed red line</u> (see also Sundararajan and Najmi, 2020). The IGs are easy to compute analytically in some cases, but for most situations, approximation is required. See (Sundararajan, Taly, & Yan, 2017) for a discussion of efficient Riemann approximation.

## 3. ANALYTICAL COMPARISONS OF BSHAP AND IG

This section discusses the connections and differences between BShap and IG for some common statistical models.

### 3.1 Special cases where BShap and IG give the same explanations

It is straightforward to verify the following result.

**Result 1: Additive models** Consider the additive model $f(x) = \sum_{i=1}^{p} f_i(x_i)$. If the $f_i$'s are all differentiable, the attribution values from BShap and IG are the same.

Hence, BShap and IG lead to the same explanations for linear, polynomial, and non-linear regression models that are additive. More generally, the equivalence holds for any generalized additive model where all components are (a.e.) differentiable.

**Result 2: Simple multiplicative models:** When $f(x) = \prod_{i=1}^{P} x_i$, BShap and IG are equivalent.

<u>Proof</u>: We can write $\phi_j^{IG}(x^O, x^R)$, the IG value for feature $j$, as

$$\phi_j^{IG}(x^O, x^R) = (x_j^O - x_j^R) \int_{\alpha=0}^{1} \prod_{i \neq j} \left( x_i^R + \alpha(x_i^O - x_i^R) \right) d\alpha,$$

where

$$\prod_{i \neq j} \left( x_i^R + \alpha(x_i^O - x_i^R) \right) = \prod_{i \neq j} \left( (1 - \alpha) x_i^R + \alpha x_i^O \right)$$
$$= \sum_{k=0}^{P-1} \sum_{S \subset P \setminus j, |S|=k} \alpha^k (1 - \alpha)^{(P-1-k)} \prod_{i \in S} x_i^O \prod_{i \notin S \cup j} x_i^R.$$

Therefore,

$$\phi_j^{IG}(x^O, x^R) = (x_j^O - x_j^R) \sum_{k=0}^{P-1} \sum_{S \subseteq P \setminus j, |S|=k} \prod_{i \in S} x_i^O \prod_{i \notin S \cup j} x_i^R \int_{\alpha=0}^{1} \alpha^k (1-\alpha)^{(P-1-k)} d\alpha$$

$$= (x_j^O - x_j^R) \sum_{k=1}^{P-1} \sum_{S \subseteq P \setminus j, |S|=k} \frac{k!\,(P-k-1)!}{P!} \prod_{i \in S} x_i^O \prod_{i \notin S \cup j} x_i^R$$

$$= (x_j^O - x_j^R) \sum_{S \subseteq P \setminus j} \frac{|S|!\,(P-|S|-1)!}{P!} \prod_{i \in S} x_i^O \prod_{i \notin S \cup j} x_i^R$$

The last equation follows from Euler integral.

On the other hand,

$$\phi_j^{SHAP}(\boldsymbol{x}^O, \boldsymbol{x}^R) = \sum_{S \subseteq P \setminus j} \frac{|S|!\,(P-|S|-1)!}{P!} \left( x_j^O \prod_{i \in S} x_i^O \prod_{i \notin S \cup j} x_i^R - x_j^R \prod_{i \in S} x_i^O \prod_{i \notin S \cup j} x_i^R \right)$$

$$= (x_j^O - x_j^R) \sum_{S \subseteq P \setminus j} \frac{|S|!\,(P-|S|-1)!}{P!} \prod_{i \in S} x_i^O \prod_{i \notin S \cup j} x_i^R.$$

Therefore, BShap and IG are equivalent.

Following Results 1 and 2, we see that the two methods give the same explanations for models of the form

$$f(\boldsymbol{x}) = \beta_0 + \sum_j \beta_j x_j + \sum_{j \neq k} \beta_{jk}\, x_j x_k + \sum_{j \neq k \neq \ell} \beta_{jk\ell}\, x_j x_k x_\ell + \cdots$$

with simple main effects, two-, three-, and higher-order interactions. The same also holds if we replace the main effects by additive models of the form $\sum_{j=1}^{p} f_j(x_j)$. But it does not hold for more general forms of interactions (see next section).

There are probably many other cases where BShap and IG give the same explanations, and we have not tried to identify all of them. Nevertheless, the above examples provide common situations where the model explanations are the same.

## 3.2 Examples where BShap and IG give different explanations

Consider first the simple case

$$f(\boldsymbol{x}) = \beta_0 + \beta_1 x_1 + \beta_2 x_2 + \beta_{12} x_1 x_2^2.$$

It is easy to show that we have

$$\phi_1^{SHAP}(\boldsymbol{x}^O, \boldsymbol{x}^R) = \beta_1(x_1^O - x_1^R) + \frac{1}{2}\beta_{12}(x_1^O - x_1^R)\left(x_2^{O\,2} + x_2^{R\,2}\right),$$

$$\phi_2^{SHAP}(\boldsymbol{x}^O, \boldsymbol{x}^R) = \beta_2(x_2^O - x_2^R) + \frac{1}{2}\beta_{12}\left(x_2^{O\,2} - x_2^{R\,2}\right)(x_1^O + x_1^R).$$

Slightly more involved computations show that

$$\phi_1^{IG}(\boldsymbol{x}^O, \boldsymbol{x}^R) = \beta_1(x_1^O - x_1^R) + \frac{1}{3}\beta_{12}(x_1^O - x_1^R)\left((x_2^O)^2 + x_2^R x_2^O + (x_2^R)^2\right)$$

$$\phi_2^{IG}(x^O, x^R) = \beta_2(x_2^O - x_2^R) + \frac{1}{3}\beta_{12}(x_2^O - x_2^R)(2(x_1^O x_2^O + x_1^R x_2^R) + x_1^O x_2^R + x_1^R x_2^O).$$

In this case, the interaction term is more complex than the form in Result 2, and the two methods give different explanation results.

Another interaction form of special interest is $f(x) = \max(x_1, x_2)$ as it arises naturally when a model is fitted using neural networks with ReLU activation. Figure 2 provides a visual display of $\phi_1^{IG}(x_1, x_2)$ and $\phi_1^{BShap}(x_1, x_2)$ for $x_1, x_2 \in [-1, 1]$. It is easy to show that $\phi_1^{IG}(x_1, x_2) = x_1$ if $x_1 + x_2 > 0$ and 0 otherwise. The expression for $\phi_1^{BShap}(x_1, x_2)$ is a bit more complex but still can be derived (but we omit the details). We see from Figure 2 that the local attributions values for the two methods can be quite different. (Sundararajan, Taly, & Yan, 2017) and (Sundararajan & Najmi, 2020) discussed this problem in the context of $\min(x_1, x_2)$, and an underlying theoretical explanation can be found in (Lundstrom, Huang, & Razaviyayn, 2022).

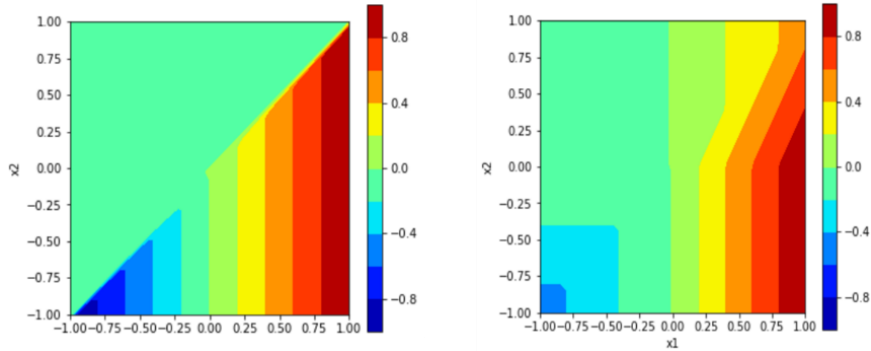

Figure 2: Attribution values for $x_1$: Left panel is $\phi_1^{IG}(x_1, x_2)$ and right panel is $\phi_1^{BShap}(x_1, x_2)$

## 4. EMPIRICAL COMPARISONS FOR TWO EXAMPLES

### 4.1 Additive Example

Consider the additive model

$$f(x) = \max(0, x_1) + x_2^3 + \exp(-2x_3) + (1 + |x_4|)^{-1} + \sqrt{|x_5|}. \qquad Eq\ (2)$$

From Result 1, we know that BShap and IG values will be the same for all five predictors. But what happens when we fit the model using a neural network with the common ReLU activation function?

We use a simulation study to examine this issue. Specifically, we let $y = f(x) + \epsilon$, where $f(x)$ is in Eq (2), and generated 10,000 simulated samples with iid Gaussian errors with variance 0.25. The predictors were sampled independently from the uniform distribution $U(-1, 2)$. We chose this uniform distribution so that the center (reference point) is not zero. We fitted feedforward neural networks (FFNNs) with ReLU activation and tuned the hyperparameters using cross validation.

We computed $\phi_i^{IG}(x)$ and $\phi_i^{BShap}(x)$ for 100 values of $x^O$ sampled randomly. We fixed $x^R$ to be the midpoint $(0.5, 0.5)$. Figure 3 shows the actual differences between BShap and IG for the fitted models for the five different predictors. The differences for $x_1, x_2$ and $x_3$ are about the same order of magnitude, while they are smaller for $x_5$ and much smaller for $x_4$. This example suggests that the differences induced by the fitted FFNN using ReLU activation are not substantial.

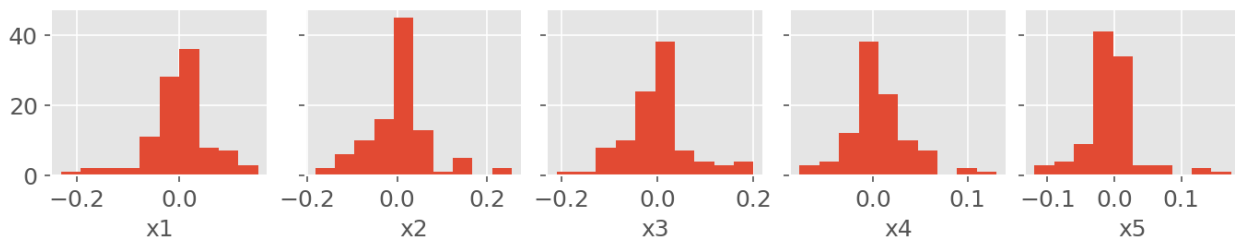

Figure 3. Histograms of actual differences between BShap and IG for the fitted models for the Additive Example

## 4.2 Example with Interactions

Consider the model

$$f(x) = x_1 + x_2 + x_3 + \cdots + x_8 + x_1 x_2 + 0.5 x_3 x_4^2 + 2 \max(x_5, x_6) + 1.5 |x_7 + x_8|. \quad Eq\ (3)$$

It includes different types of interactions that are common. Figures 4 shows the actual differences in the attributions given by BShap and IG for selected predictors: $x_3, x_4, x_5, x_7$. Behaviors for the other variables can be inferred from symmetry. Note from Result 2 that, for the true model, BShap and IG attributions should be the same.

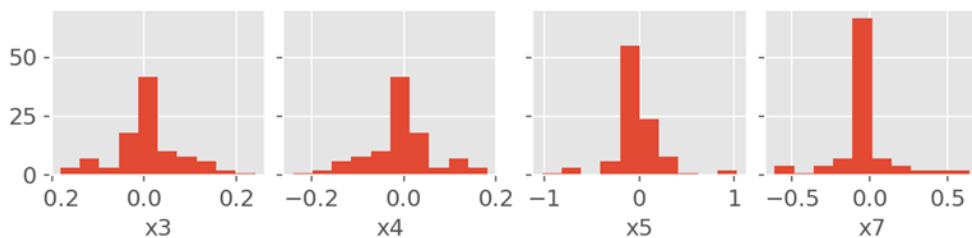

Figure 4. Histograms of the actual differences between BShap and IG for the Interaction Example

Again, we consider their behavior when we fit the model using a neural network with the ReLU activation function. Specifically, we let $y = f(x) + \epsilon$, where $f(x)$ is in Eq (3). The rest of the simulation details are the same as in Section 4.1. Figure 5 shows the actual differences between BShap and IG for the fitted models for the five different predictors.

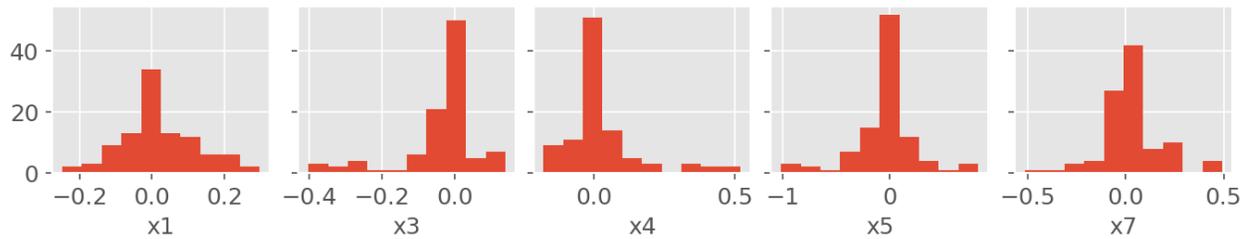

Figure 5. Histograms of actual differences between BShap and IG for the fitted models for the Interaction Example

Recall that the difference for the true model is zero, the observed differences for $x_1$ in Figure 5 arise from the fitted model. Larger differences can be observed for other predictors. However, a comparison with Figure 4 shows these differences after fitting the model are not substantially larger.

## 5. Discussion

This paper provides some additional insights into the behavior of two local attribution methods: BShap and IG. Their axiomatic properties have been developed in the literature and summarized in the Appendix. Both of them are based on path integrals, and other path-based methods have been discussed in the literature for image and natural language processing related tasks (Kapishnikov, Bolukbasi, Viegas, & Terry, 2019; Kapishnikov, et al., 2021; Sanyal & Ren, 2021; Sikdar, Bhattacharya, & Heese, 2021).

We considered common classes of models where BShap and IG give the same explanation and where they are different. Our limited simulations show that the differences are not excessively big. Therefore, on balance, their choice is based on convenience and applicability. The use of IG with tabular data requires smooth (differentiable) response surfaces, and hence they are not applicable to tree-based algorithms. There is an attempt (Merrill, Ward, Kamkar, Budzik, & Merrill, 2019) to generalize IG to situations with jumps. But the approach just uses BShap at those points, so it is not particularly novel.

Given the widespread nature of tree-based algorithms, we propose the use of BShap as a more generally applicable method.

# 7. APPENDIX

## 7.1. PROPERTIES OF BSHAP AND IG

Sunderarajan and Najmi (2020) describe five axiomatic properties that are shared by BShap and IG. We summarize them here for completeness.

**Definition (Efficiency).** For every point $x^O$, and baseline $x^R$, the attributions add up to the difference $f(x^O) - f(x^R)$ for the baseline approach.

**Definition (Linearity).** The attributions of the linear combination of two functions $f_1$ and $f_2$ is the linear combination of the attributions for each of the two functions.

**Definition (Dummy).** Dummy features get zero attributions. A feature $i$ is dummy in a function $f$ if for any two values $x_i^O$ and $x_i^R$, and every value $x_{N \setminus i}$ of the other features, $f(x_i^O; x_{N \setminus i}) = f(x_i^R; x_{N \setminus i})$.

**Definition (Affine Scale Invariance).** The attributions are invariant under a simultaneous affine transformation of the function and the features.

**Definition (Symmetry).** For every function $f$ that is symmetric in two variables $i$ and $j$, if the $x^O$ and $x^R$ are such that $x_i^O = x_j^O$ and $x_i^R = x_j^R$, then the attributions for $i$ and $j$ should be equal.

Demand monotonicity is a unique feature of BShap defined as below:

**Definition (Demand Monotonicity).** For every feature $i$, and function $f$ that is non-decreasing in $i$, the attribution of feature $i$ should only increase if the value of feature $i$ increases, with all else held fixed. This simply means that if the function is monotone in a feature, that feature's attribution should only increase if the point of interest's value for that feature increases.

IG satisfies the proportionality axiom that is desirable in computer vision.

**Definition (Proportionality).** If the function $f$ can be rewritten as a function of $\sum_i x_i^O$, and the baseline $x^R$ is zero, then the attributions are proportional to the values $x$.

## 7.2. Choice of Baseline or Reference Point

There is an extensive discussion in the literature on the choice of a baseline or reference point when making attributions. (Sundararajan, Taly, & Yan, 2017; Sturmfels, Lundberg, & Lee, 2020; Haug, Zurn, El-Jiz, & Kasneci, 2021) note that the concept of baseline is related to the concept of missingness, i.e., what value would a feature take if it is considered missing. Using a zero or the mean vector as the baseline is not always a good idea. In our view, the choice must be dictated by the application. See Nair et al. (2022) for a discussion on the use of difference reference points in the context of adverse actions for credit decisions.